%% file: acl_latex.tex
\pdfoutput=1
\documentclass[11pt]{article}

\usepackage{latex/acl}

\usepackage{times}
\usepackage{latexsym}

\usepackage[T1]{fontenc}
\usepackage[utf8]{inputenc}

\usepackage{microtype}

\usepackage{inconsolata}

\usepackage{amsmath}
\usepackage{subfigure}
\usepackage{caption}
\usepackage{multirow}
\usepackage{todonotes}
\usepackage{cleveref}
\usepackage{float}
\usepackage{bbding}

\crefname{section}{§}{§§}
\Crefname{section}{§}{§§}

\usepackage{amsthm,amsmath,amssymb}
\usepackage{mathrsfs}
\usepackage{mathtools}
\newcommand{\abs}[1]{\lvert#1\rvert}

\usepackage{amsmath}

\title{LLMs as Narcissistic Evaluators: When Ego Inflates Evaluation Scores}

\author{ 
{ \bf {Yiqi Liu}}$^1$ $\-$ $\-$ $\-$ {\bf {Nafise Sadat Moosavi}}$^2$ $\-$ $\-$ $\-$ {\bf {Chenghua Lin}}$^1$ \\
{$^{1}${University of Manchester} $\-$ $\-$ $\-$ $^2${University of Sheffield}} \\
\texttt{{yiqi.liu-6@postgrad.manchester.ac.uk}}$\-$ $\-$ $\-$ $\-$  \texttt{{n.s.moosavi@sheffield.ac.uk}}$\-$ $\-$ $\-$ $\-$  \\
 \texttt{{chenghua.lin@manchester.ac.uk}}
}

\begin{document}

\maketitle

\begin{abstract}
Automatic evaluation of generated textual content presents an ongoing challenge within the field of NLP. Given the impressive capabilities of modern language models (LMs) across diverse NLP tasks, there is a growing trend to employ these models in creating innovative evaluation metrics for automated assessment of generation tasks. This paper investigates a pivotal question: \textit{Do language model-driven evaluation metrics inherently exhibit bias favoring texts generated by the same underlying language model?} Specifically, we assess whether prominent LM-based evaluation metrics (e.g. BARTScore, T5Score, and GPTScore) demonstrate a favorable bias toward their respective underlying LMs in the context of summarization tasks. Our findings unveil a latent bias, particularly pronounced when such evaluation metrics are used in a reference-free manner without leveraging gold summaries. These results underscore that assessments provided by generative evaluation models can be influenced by factors beyond the inherent text quality, highlighting the necessity of developing more reliable  evaluation protocols in the future. %Additionally, we offer insights into reducing the identified bias when utilizing existing LM-based evaluators.
\end{abstract}

\section{Introduction}
Evaluation is a fundamental element in both tracking progress and ensuring meaningful advancements across various dimensions within the field of Natural Language Processing.  Therefore, the reliability of evaluation metrics plays a critical role in this process. Evaluating generated texts is one of the challenging and open problems in NLP given that different forms can convey the same meaning. This challenge has led to the development of various evaluation metrics for tasks involving Natural Language Generation (NLG). While human evaluation by experts stands as the most reliable approach for assessing generated outputs, it is costly and time-consuming, limiting its broader use. As a result, automatic evaluation metrics have emerged as practical alternatives to keep pace with the rapid progress in NLP \cite{van-der-lee-etal-2019-best}. Recent evaluation metrics for generation tasks, such as BERTScore \cite{Zhang2019-gi}, BARTScore \cite{Yuan2021-id}, T5Score \cite{qin2022t5score}, GPTScore \cite{fu2023gptscore}, and G-Eval \cite{liu2023geval}, increasingly rely on pretrained language models.
\begin{figure}[t]
    \centering
    \includegraphics[width=1\linewidth]{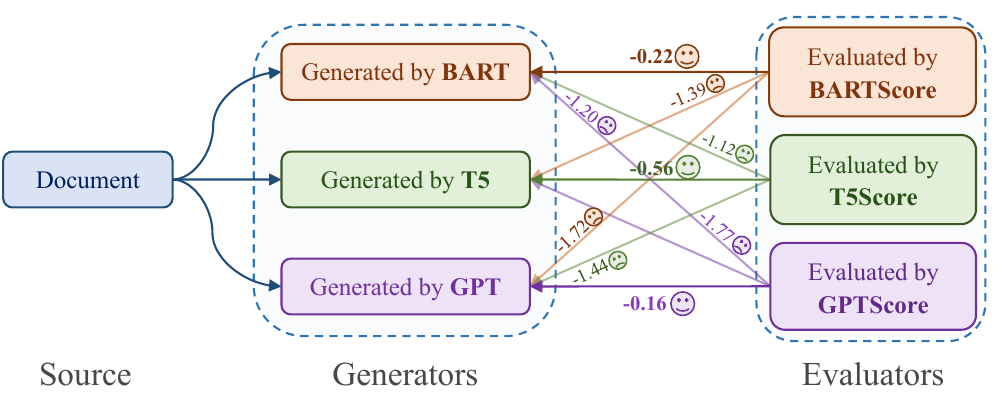}
    \caption{Examining the inherent bias within generative evaluation metrics towards outputs created by their underlying model reveals a clear existence of this bias. Our analysis shows that these metrics tend to assign inflated scores to outputs generated by the very model they are based on.}
    \label{fig:enter-label}
\end{figure}
However, this trend poses a paradox, as the very outputs being evaluated are generated by these pretrained language models, raising concerns about inherent biases. For instance, an evaluation metric based on the BART model might yield inflated scores for outputs produced by a BART-based language model. 

In this paper, we systematically investigate this potential bias, utilizing six prominent language models, namely BART \cite{lewis-etal-2020-bart}, T5 \cite{raffel2020exploring}, GPT-2 \cite{radford2019language}, GPT-3 \cite{Brown2020-qj},  FLAN-T5 \cite{chung2022scaling}, and Cohere along with their corresponding evaluation metrics (e.g. BARTScore, T5Score, and GPTScore) or conditional generative probability, for the task of summarization, which is a typical task in natural language generations and frequently employed in automatic text evaluation. Our analysis involved examining numerous variations of these six families of generative models, considering their varying sizes and finetuning settings both as generators and evaluators.
%\textcolor{red}{[CL: we need to provide some (brief) justification why focusing on sumamrisation]}

We conducted our analysis using the CNN/Daily Mail \cite{hermann2015teaching} and XSUM \cite{narayan2018don} summarization datasets. The assessment covers two settings: reference-based, using gold summaries for evaluation (a common approach in supervised summarization), and reference-free, comparing generated summaries against source documents (a common approach in both unsupervised summarization and factuality assessment).

Based on our analysis, we have derived the following findings: (1) Generative evaluators tend to assign higher scores to the content generated by the same underlying model. This bias becomes more pronounced when the fine-tuning configuration and model size match for both the generator and evaluator. (2) Inflated scores are particularly noticeable in the reference-free setting, which is concerning due to the popularity of this evaluation approach for assessing the factual correctness of generated texts \citep{koh-etal-2022-far}. (3) Apart from self-bias, inflated scores are also influenced by the preference for longer summaries by certain evaluators.

Our work has implications for model selection, evaluation strategies, and the development of more reliable and unbiased evaluation metrics in the field of natural language generation.

\section{Related Work}

\paragraph{Reference-based Evaluation Metrics} 

Reference-based metrics are commonly used to evaluate text generation tasks, including summarization, by measuring the similarity between generated and reference texts. Traditionally, metrics like BLEU \cite{papineni-etal-2002-bleu} and ROUGE \cite{lin-2004-rouge} were employed to assess a generated text based on surface-level similarities, measured through the n-gram overlap between the generated and reference texts.

Recent trends in summarization evaluation lean towards semantic-level assessments, moving beyond direct word overlap comparisons. Notable metrics embracing this approach include BERTScore \cite{Zhang2019-gi}, MoverScore \cite{zhao-etal-2019-moverscore}, BARTScore \cite{Yuan2021-id}, BLEURT \cite{sellam-etal-2020-bleurt}, and variations thereof. By leveraging pretrained language models, these metrics focus on capturing semantic content, providing a more nuanced and accurate evaluation of summarization system outputs.

\paragraph{Reference-free Evaluation Metrics} 
With the widespread use of generation models across diverse domains, the need for reference-free evaluation metrics has surged. In response to this challenge, recent attention has been directed towards metrics that enable the evaluation of generated texts solely based on source documents, especially when annotated reference texts may not be available for new domains \cite{bohm-etal-2019-better, gao-etal-2020-supert, wu-etal-2020-unsupervised, chen-etal-2021-training, scialom-etal-2021-questeval, honovich-etal-2021-q2, zhong-etal-2022-towards, liu2023geval}.

Representative reference-free metrics in recent years include generative evaluation models, exemplified by BARTScore \cite{Yuan2021-id} and GPTScore \cite{fu2023gptscore}, which are also used for reference-based evaluation. These metrics frame text evaluation as a natural language generation task, intuitively assigning higher probabilities to higher-quality generated texts. For instance, a recent study by \citet{koh-etal-2022-far} has acknowledged BARTScore in reference-free mode as the factual consistency metric with the highest overall correlation to human factual consistency scores, particularly in the context of long document abstractive summarization. Therefore, the reliability of these metrics is important given their use for evaluating sensitive aspects such as factuality correctness.

\paragraph{Automatic Evaluation Metrics Pitfalls}
Despite their widespread use, automatic evaluation metrics have notable shortcomings. These metrics may not be robust when faced with challenges such as spurious correlations, noise, or out-of-domain texts \citep{sai-etal-2021-perturbation, vu-etal-2022-layer, durmus-etal-2022-spurious, zhao-etal-2023-evaluating,he-etal-2023-blind}. Furthermore, their effectiveness diminishes when evaluating very long documents \citep{Amplayo2022SMARTSA}. There is also evidence suggesting a potential bias towards ranking extractive summaries higher than abstractive ones \cite{Amplayo2022SMARTSA}.

Traditional reference-based evaluation metrics such as ROUGE or BLEU have been criticized for their inability to measure content quality or capture syntactic errors \cite{reiter-belz-2009-investigation}. 
Consequently, these traditional metrics often exhibit weak correlations with human judgements, demonstrating that they cannot accurately reflect the real-world performance of generation systems~\cite{peyrard-2019-studying, mathur-etal-2020-tangled}. 
% These metrics may inaccurately reflect the reliability of generation systems. 
For example, they might assign high scores to outputs that are fluent but meaningless and unfaithful, as long as many of the same words are used \cite{gehrmann-etal-2021-gem}. Although embedding-based metrics (e.g., BERTScore) show improved performance in similarity measurement, they are still inadequate for assessing the extent of shared information between two summaries, a crucial indicator of summary information quality \cite{deutsch-roth-2021-understanding}.

Reference-free metrics, on the other hand, exhibit a bias towards outputs generated by models that are more similar to their own \cite{deutsch-etal-2022-limitations}. To the best of our knowledge, this study represents the initial attempt to perform an exploration, which has not yet been undertaken systematically. Additionally, question-answering-based reference-free metrics for summarization evaluation are prone to inheriting errors within summaries \cite{kamoi2023shortcomings}.
Metrics based on Large Language Models, which are capable of conducting both reference-based and reference-free evaluations, typically demonstrate superior correlations with human quality judgements across diverse NLG tasks and evaluation dimensions \cite{deutsch-etal-2022-limitations}. 
While prior work has reported that LLM-based metrics prefer LLM-generated text, raising a concern about the shortcomings of LLMs as evaluators \cite{liu2023geval}, our work conducts a systematic evaluation to address a fundamental question: \textit{Do language model-driven evaluation metrics inherently display bias favouring texts generated by the same underlying language model?} We explore this question across both reference-based and reference-free evaluations and for a range of different large language models.
% \textcolor{blue}{please mention some of those shortcomings(\checkmark)}

%However, automatic evaluation metrics have some pitfalls for their own reasons. For instance, they are sensitive to the inputs, a slightly changes could cause a significant changes to the output. Also, 
%Some researchers have reported instances of such bias along with their works that Generative Pre-Trained (GPT) based evaluators favoring content generated by GPT variants \cite{liu2023geval, wang2023far}.

%\todo[inline]{review existing studies on the drawbacks of automatic evaluation metrics, and then at the end mention that the bias that you are investigating is in that line of research and it hasn't been investigate thoroughly before}
%

%Various of methods were proposed to address this issue, such as overlap-based \textsc{ROUGE} criterion \cite{lin-2004-rouge}, and embedding-based \textsc{BERTScore} \cite{Zhang2019-gi}, which focusing on word-level semantic similarity. 

% \newline
% \newline
% \textbf{Generative Evaluation}  

\section{Methodology}
To investigate the impact of the model's self-bias---determining whether a language model-based evaluator favours outputs generated by a similar language model---we conduct a comprehensive series of experiments involving both quantitative comparisons and qualitative analysis.
Our quantitative comparisons involve using language models of varying sizes and finetuning configurations as both the evaluator and generator models. This structured approach enables us to systematically examine the potential bias across different LM configurations. 
Subsequently, we verify the results through qualitative analysis using a subset of models' summaries that are accompanied by human evaluation to further demonstrate that higher scores produced by evaluators as a result of self-bias do not necessarily correlate with higher quality generated outputs. 
 
%Our study involves bias analysis and meta-evaluation. The former is a comparative investigation that employs generative evaluators to systematically assess summaries from different generation models. The latter utilizes human annotation data to identify evaluators that exhibits higher correlation with human judgment and lower bias.

\subsection{Evaluators}

%Demands for automatic evaluation on text generated by PLMs is surging along with the rapid development of PLMs themselves. The goal of automatic text evaluation is to make an assessment to the content generated by language models. 
%In this paper, we 
We describe the evaluation process as follows: given a \textit{source} text \textbf{\emph{s}}, a human written \textit{reference} \textbf{\emph{r}}, generate a \textit{hypothesis} \textbf{\emph{h}}, which can be represented as:
\begin{equation}
    y = f(\textbf{\emph{h}},a,\mathcal{S})
\end{equation}
where \textbf{\emph{h}} denotes hypothesis, \textit{a} refers to the aspect to evaluate, and \begin{math}\mathcal{S}\end{math} denotes supplementary text (i.e., \textbf{\emph{s}} or \textbf{\emph{r}}) that is employed alongside evaluations in various settings \cite{fu2023gptscore}. For instance, it could be the source text \textbf{\emph{s}} in a \textit{reference-free} scenario which assesses the summary based on the source article directly \cite{fabbri-etal-2021-summeval}. Whereas in the \textit{reference-based} paradigm, the evaluation considers semantic overlap between the generated hypothesis \textbf{\emph{h}} (e.g. model generated summaries) and reference summaries \textbf{\emph{r}}~\cite{bhandari-etal-2020-evaluating}.

% Three 
The evaluators (i.e. based on BART, T5, GPT model variants as well as Cohere) utilised in our study all share a conditional probability paradigm, which can generally be formulated as
\begin{equation}\label{eq.2}
        \!\!\!\!\!\!\!\!\mathrm{Score}(\textbf{\textit{h}} | d,a,\mathcal{S}) = \sum_{t=1}^{m}w_{t}\log{p}(h_{t} | \textbf{\textit{h}}_{<t},\mathcal{S},\theta). 
\end{equation}
% \textcolor{blue}{what is d in the equation? (\checkmark)}
Here \begin{math}\theta\end{math} is the model parameter, \begin{math} d \end{math} refers to the task description and \begin{math} w_{t} \end{math} denotes the weight of the token at position \begin{math} t \end{math}, where previous works normally treat each token equally \cite{Yuan2021-id, fu2023gptscore}. We provide further descriptions of each type of evaluator below.

%Under a sequence-to-sequence model\footnote{\textcolor{orange}{The models used in this paper include BART, T5, GPT, Cohere and their variants.} } parameterised by \begin{math}\theta\end{math},  where \begin{math} d \end{math} refers to the task description and \begin{math} w_{t} \end{math} denotes the weight of the token at position \begin{math} t \end{math}. Previous works normally treat each token equally \cite{Yuan2021-id, fu2023gptscore}.
%; Finally, \begin{math} \mathcal{S} \end{math} is determined according to the aspect to evaluate.

\paragraph{BARTScore}
BARTScore \cite{Yuan2021-id} introduced the generative evaluation approach treating text assessment as a generation task, employing probability of the text being generated by BART-based models \cite{lewis-etal-2020-bart} to assess the quality of text generated  
%leveraging open-source pre-trained sequence-to-sequence models as evaluators 
across various tasks such as machine translation, summarization, and data-to-text. 

\paragraph{T5Score}
T5Score \cite{qin2022t5score} was proposed providing both generative and discriminative training strategies for assessing T5-variant models as the core of this generative evaluation paradigm\footnote{In our work, the training process of T5Score models only involves generative training due to the unavailability of publicly accessible checkpoints trained in a discriminative manner.}. The integration of dual training strategies enables more types of data to be incorporated into the metric. T5Score closely aligns with BARTScore in terms of evaluation framework. Thus, when only considering the generative training strategy, T5Score is analogous to BARTScore, but for the T5 model series.

\paragraph{GPTScore}
 Leveraging generative models to conduct evaluation has been further advanced with various of more recent large language models \cite{fu2023gptscore}, showing a great performance and covering a rich variety of aspects for comprehensive evaluations. With an understanding of natural language instructions, GPTScore (including GPT-X and FLAN-T5 models) can perform intricate and personalized assessments without additional training.

\paragraph{Cohere} We additionally include Cohere, the more recent language model to enrich our assessments.  
%\textcolor{blue}{named command provided by Cohere \footnote{\url{https://docs.cohere.com/reference/generate}}, as the flagship model with the excellent performance on generation task. 
The evaluation scores assigned by the model is calculated according to Eq.~\ref{eq.2}, aligned with BARTScore, T5Score, and GPTScore.
 
% \textcolor{blue}{what is the evaluation equation? Is using structures and prompts the only difference? Please elaborate(\checkmark)}\textcolor{green}{YL: The evaluation of GPTScore can also be represented by Eq.2. To a certain extent, differences between BARTScore and GPTScore are majorly structures and prompts. (Since they are proposed by the same research team)}
\subsection{Generation Models}

We analyze different variants of the BART, T5, GPT-2, GPT-3, FLAN-T5 and Cohere models, taking into account two different variables: the \textit{model size} and the \textit{finetuning dataset}. Regarding size, we consider small, base, medium, and large variations of each model, when available. For the finetuning dataset, we examine three distinct settings: (1) using the pretrained language model without finetuning on a summarization dataset, (2) finetuning on CNN, and (3) finetuning on XSUM. For instance, BART-Base-CNN represents a BART-base model that is finetuned on the CNN dataset. 
For each of the model types, we have used their corresponding standard prompts for the task of summarization.\footnote{More details about the corresponding summarization prompts are included in Appendix~\ref{A.2}.} 

To ensure the reproducibility of our analysis, we exclusively employ publicly available checkpoints for the utilized models. 
Apart from the GPT3-Curie model that is taken from the OpenAI API
%\footnote{Note that the newer models offered by the OpenAI API, such as GPT-4, no longer support providing token probabilities, and hence they cannot be tested in our experiment.} 
and generation model obtained from Cohere,  %\footnote{\textcolor{red}{We did not utilize summarize model directly from Cohere because acquiring token probabilities is only supported by its generation models}}
the rest of the models are taken from the Hugging Face model hub\footnote{\url{https://huggingface.co/models}.}.

We use each of these generation models both for generating the summaries\footnote{We use the zero-shot setting for the models that are not finetuned on summarization datasets.} as well as the underlying model for the LM-based evaluator. 
All the checkpoints used for generators and evaluators in our experiments can be found in the Appendix \ref{A} (Table \ref{tabs:name.gene} and Table \ref{tabs:name.eval}).  
%For the evaluator, \textcolor{orange}{the BART and T5 variants are employed for evaluation using the BARTScore code} \textcolor{blue}{and T5Score code respectively},\footnote{Please note that under unsupervised training strategy in T5Score, the evaluation framework of T5Score and BARTScore are identical.} \textcolor{orange}{the GPT and FLAN-T5 variants are utilized using the GPTScore code.} \textcolor{blue}{Cohere model for evaluation are implemented according to Eq. \ref{eq.2}.}
% \textcolor{blue}{Are they different for reference-based? how different they are? you previously mentioned that they are identical for the generative setting. This should also be explained with that. (\checkmark)}\textcolor{green}{ YL: Yes it should be talked in terms of different training strategy, and I have corrected it. }  

%Generation models, including diverse sizes and fine-tuning setups of BART, T5, and GPT variants, produce summaries for source texts from CNN/DM and XSUM datasets. We use pipeline module from Hugging Face and OpenAI API as our generation framework. All generation models are publicly accessible. Since not all above-mentioned models are able to generate summarizations directly, inputs for certain models utilize prefixes or suffixes to enable summarization. 

%\subsection{Bias Experiment}
%The fundamental process of bias analysis entails using generators to produce summaries from source texts in established summarization datasets, extracting summaries from human-annotated datasets, and evaluating these summaries using BARTScore, T5Score, and GPTScore frameworks.

\subsection{Datasets} 
\label{dataset}
We use documents from two well-established summarization datasets including CNN/DailyMail \cite{hermann2015teaching,nallapati2016abstractive} and the extreme summarization (XSUM) dataset \cite{narayan2018don}. 

For quantitative comparisons, we randomly selected 500 documents from each of these datasets. We provide these documents to each of the generation models to obtain their corresponding generated summaries. 
For qualitative analysis, we use the SummEval benchmark \cite{fabbri-etal-2021-summeval} and the RoSE benchmark \cite{liu2022revisiting}. These benchmarks include summaries from various generation models, as well as human evaluations, enabling us to assess the quality of these summaries.

The SummEval benchmark contains summaries generated by various summarization models (i.e. BART, T5 and GPT2) for 100 articles from the CNN/DM test set, with each summary supplemented by human annotations. More specifically, SummEval incorporates human annotations by both expert and crowd-sourced human annotators, targeting dimensions of coherence, consistency, fluency, and relevance. Ratings are on a scale of 0 to 5, with higher values indicating better performance.

Similarly, RoSE contains summaries generated by recent generative models based on CNN/DM documents, accompanied by their corresponding human evaluations. We use 100 summaries from each of the BART and GPT-3 models from the ROSE benchmark. The RoSE benchmark proposed an assessment protocol termed ``Atomic Content Units'' (ACUs) \cite{liu2022revisiting}. ACU score gauges quality of evaluated summaries based on whether the presence of single facts (i.e., atomic facts) from reference are included in the evaluated summaries. ACU score is calculated by ACU matching:
% \textcolor{blue}{do you mean summaries? (\checkmark) }
\begin{equation}{
    f(s,\mathcal{A})=\frac{\abs{\mathcal{A}_s}}{\abs{\mathcal{A}}}
}\end{equation}\label{eq.acu}
where \begin{math}\mathcal{A} \end{math} is a set of ACUs from gold summaries and \begin{math}\mathcal{A}_s\end{math} denotes the ACUs of candidate summary \begin{math}\textit{s}\end{math}.
%\begin{math}\textit{s}\end{math} is the candidate summary. 
%\textcolor{orange}{Thus, \begin{math}\mathcal{A}_s\end{math} denotes ACUs from candidate summary.}

\input{tables/table_annotation}
The distribution of human scores in RoSE and SummEval are given in Table~\ref{tab:human_anno}.

% \textcolor{blue}{Set of ACUs in what? gold summary or the source?(\checkmark) }\textcolor{green}{YL: ACU scores that RoSE provided all consider gold summary, I have made the definition of ACU clearer.} 

% \footnote{More details on these benchmarks are provided in Appendix \ref{A.1}.}

% It's important to note that SummEval and RoSE employ different types of human evaluations.

%The first type comprises a substantial amount of source texts drawn from well-established summarization datasets, namely CNN/DailyMail \cite{hermann2015teaching} and the extreme summarization (XSUM) dataset \cite{narayan2018don}. From each dataset, we randomly selected 500 texts from each of these datasets to serve as source material for the subsequent generation process. The second type involves two distinct datasets to facilitate comprehensive analysis. The first dataset contains summaries generated by various summarization models for 100 articles from the CNN/DailyMail test set, supplemented with human annotations (SummEval; \cite{fabbri-etal-2021-summeval}); the second dataset with summary-level annotations whose summaries derived from recent generative models including GPT-3 (RoSE; \cite{liu2022revisiting}). Further details are provided in Appendix \ref{A.1}.
% Source texts are chosen from the mainstream summarization datasets CNN/DailyMail \cite{hermann2015teaching} and the extreme summarization (XSUM) dataset \cite{narayan2018don}. From each dataset, we randomly sampled at a quantity of 500 texts as source material being summarized by generators in the subsequent process.

\begin{figure*}[htbp]
\centering
\includegraphics[width=0.9\linewidth]{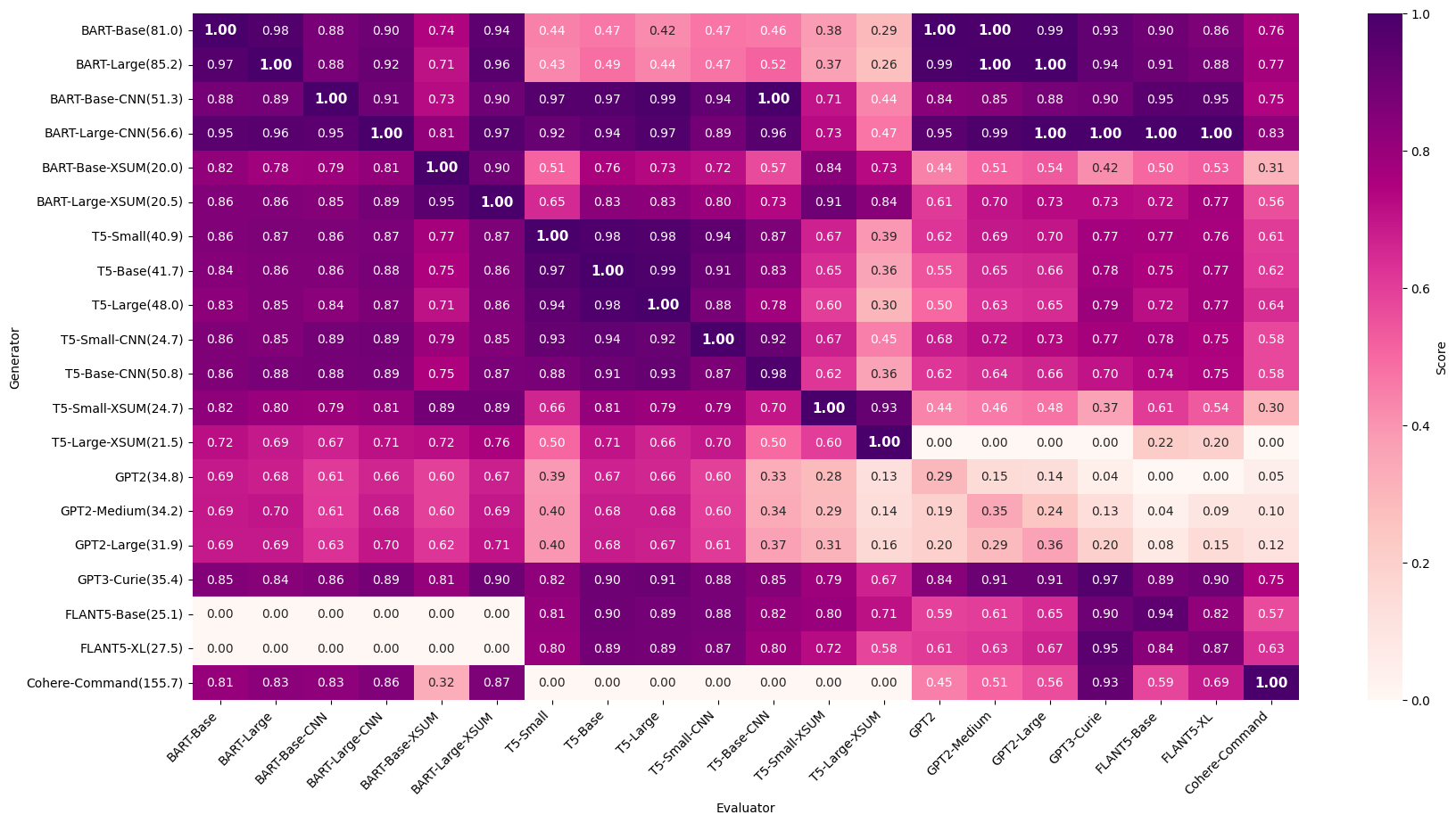}
\caption{Assessing Bias in the CNN/DM Dataset using heatmaps in the \textit{reference-free} setting. Observing the darkest cells along the diagonal line, from the top left to the bottom right, indicates a distinct bias among evaluators towards their respective models. All evaluator scores are normalized to a range between 0 and 1. Additionally, the number in the bracket represents the average length of summaries (measured in words) produced by the respective model.}
\label{Heat.cnn}
\end{figure*}

\subsection{Quantitative Comparisons}  
We employ 20
% 21 
language model-based evaluators for our experiments including six \textsc{BARTScore} evaluators \cite{Yuan2021-id}, seven \textsc{T5Score}  evaluators \cite{qin2022t5score}, six GPTScore evaluators, and the Cohere evaluator.\footnote{Appendix \ref{A.2} contains more details about the evaluators.} 

% Summaries generated by 8 different configurations across the above mentioned 3 dataset are selected for our analysis. The corresponding generators containing T5-XXL \cite{raffel2020exploring}, and T5-BASE are chosen in \textsc{seahorse}, T5-BASE, GPT-2 \cite{ziegler2019fine}, and BART \cite{lewis-etal-2020-bart} in SummEval, along with BART variants and GPT-3 \cite{Brown2020-qj} in RoSE. Details about their fintuning configurations can be seen in Appendix \ref{A.2}.

% We employ 7 LLM-based evaluators for experimentation, consisting of 3 \textsc{BARTScore} \cite{Yuan2021-id}, and three T5-based generative evaluators, alongside a GPT-2 based evaluator. The BARTScore variants include: Vanilla BART, BART fine-tuned with the CNN/Daily Mail (CNN/DM) dataset, and BART fine-tuned with the Extreme Summarization (XSUM) dataset \cite{narayan2018don}. Similarly, the T5-based generative evaluators encompass: T5-base, T5 fine-tuned with CNN/DM, and T5 fine-tuned with XSUM. More details of model can be seen in Appendix \ref{A.3}.
We assess the evaluators in two settings: (a) reference-free, where the metric evaluates the likelihood of the summary being generated from the source text, and (b) reference-based, where the generated summary is evaluated based on the reference summary.
%The reference-free metrics are majorly taken into consideration in this paper. In a reference-free setting, the metric evaluates the likelihood of the hypothesis being generated solely from the source text, which focuses on \textit{Faithfulness} (\textbf{\emph{s}} $\rightarrow$ \textbf{\emph{h}}) \cite{Yuan2021-id}. In our study, we consider each source text and its corresponding generated summary as an input pair for the evaluator.

Due to the nature of log probabilities, original scores from each evaluator is be \textit{negative}, and a higher score indicates better quality according to the evaluator. When weights $w_{t}$ in Eq.~\ref{eq.2} are treated equally, the evaluation protocols of BARTScore, T5Score, and GPTScore are all conditional probability paradigms.
% \textcolor{blue}{I don't understand the previous sentence(\checkmark)}\textcolor{green}{YL: According to Eq.2, generative evaluation scores can be regarded as weighted sum of probabilities, $w_{t}$ can be tf-idf or other things, previous papers normally each $w_{t}$ the same, which means each probability have the same weight.} 
To ensure comparability among the scores provided by 20 distinguished evaluators, a uniform normalization process is applied to the scores generated by each evaluator. The normalization procedure standardizes the scores across a scale ranging from 0 to $\alpha$ \footnote{In our work, we set parameter $\alpha$ to 1.} as formulated in Eq.~\ref{eq.4}, where $X_{i,j}$ indicates scores evaluated by the $j$-th evaluator on summaries generated by the $i$-th generator. 
\begin{equation}
    X_{i,j}^{norm}=\frac{\alpha (X_{i,j}-\min_{i}X_{i,j})}{\max_{i}X_{i,j}-\min_{i}X_{i,j}}
    \label{eq.4}
\end{equation}
In this context, a normalized score of $\alpha$ signifies the highest quality attributed by the evaluator, while a score of 0 indicates the lowest quality.
% \textcolor{blue}{Could you please include the normalization equation here? or in the appendix, if you don't have space(\checkmark)}\textcolor{green}{YL: Equation and explanations about variables are added below.}

As the length of the generated summary is a key factor influencing the evaluation results, we further analyse the impact of lengths for the content generated by the models along with the experiments. In this regard, we also compute the correlations between the length of the text and the scores assigned by evaluators to identify trends in evaluators' preferences.
%\textcolor{orange}{Furthermore, \textcolor{red}{biases of evaluators towards specific lengths are inevitable}, as this is a key factor influencing the results of our quantitative experiments. We analyse the impact of lengths for the content generated by the models along with the experiments. We also compute the correlations between the length of the text and the scores assigned by evaluators to identify trends in evaluators' preferences.}

% Reference-based setting we choose in our work is \textit{Recall} (\textbf{\emph{h}} $\rightarrow$ \textbf{\emph{r}}), which mainly consider semantic coverage between generated hypothesis and golden reference.

\subsection{Qualitative Analysis} \label{qualitative analysis}
For qualitative analysis, we employ Spearman Correlations \cite{ZarSpearman} and Kendall Correlations \cite{freedman2007statistics}, which respectively assess monotonic relationships and order associations between human evaluations and LM evaluator scores. They are common metrics for assessing correlations with human judgements.

%Human annotated data from SummEval and RoSE are correlated with the scores provided by evaluators in bias experiment. 
For the SummEval dataset, we calculate the correlations for four aspects (i.e. \textit{Coherence}, \textit{Consistency}, \textit{Fluency} and \textit{Relevance} ), aligned with the reference-free input setting in the evaluation protocol as specified by \citet{Yuan2021-id}. For the evaluations based on the RoSE benchmark, we use ACU annotations that are suited for reference-based summary salience evaluation. Therefore, we employ the correlation values obtained from the SummEval dataset for the reference-free setting and those from the RoSE benchmark for the reference-based setting.

% In our investigation, we utilize 3 distinct datasets to ensure the comprehensive analysis: a collection of summaries generated by various summarization models on 100 articles from CNN/DailyMail\cite{hermann2015teaching} test set, incorporated with human annotations (\textbf{SummEval}; \cite{fabbri-etal-2021-summeval}); the dataset with summary-level annotations whose summaries derived from recent generative models including GPT-3 (\textbf{RoSE}; \cite{liu2022revisiting}). Details are available in Appendix \ref{A.1}.
\section{Experimental Results}
\label{sec_results}
% \subsection{Bias Analysis}
% Table \ref{tab:sh_score} shows the average scores under reference-free setting (i.e., \textit{source} $\rightarrow$ \textit{hypothesis}). Summaries evaluated by evaluators with similar structure are italicized, those assessed by evaluators fine-tuned on identical datasets are underlined, and the highest score assigned for each generator is marked in bold.
% Table \ref{tab:hr_score}
% \subsection{Human Evaluations}

\subsection{Quantitative Comparisons: Assessing Self-Bias in LM-Evaluators Towards Their Own Output}
\label{unsupervised-experiments}

\begin{figure*}[tb]
\centering
\includegraphics[width=1.0\linewidth]{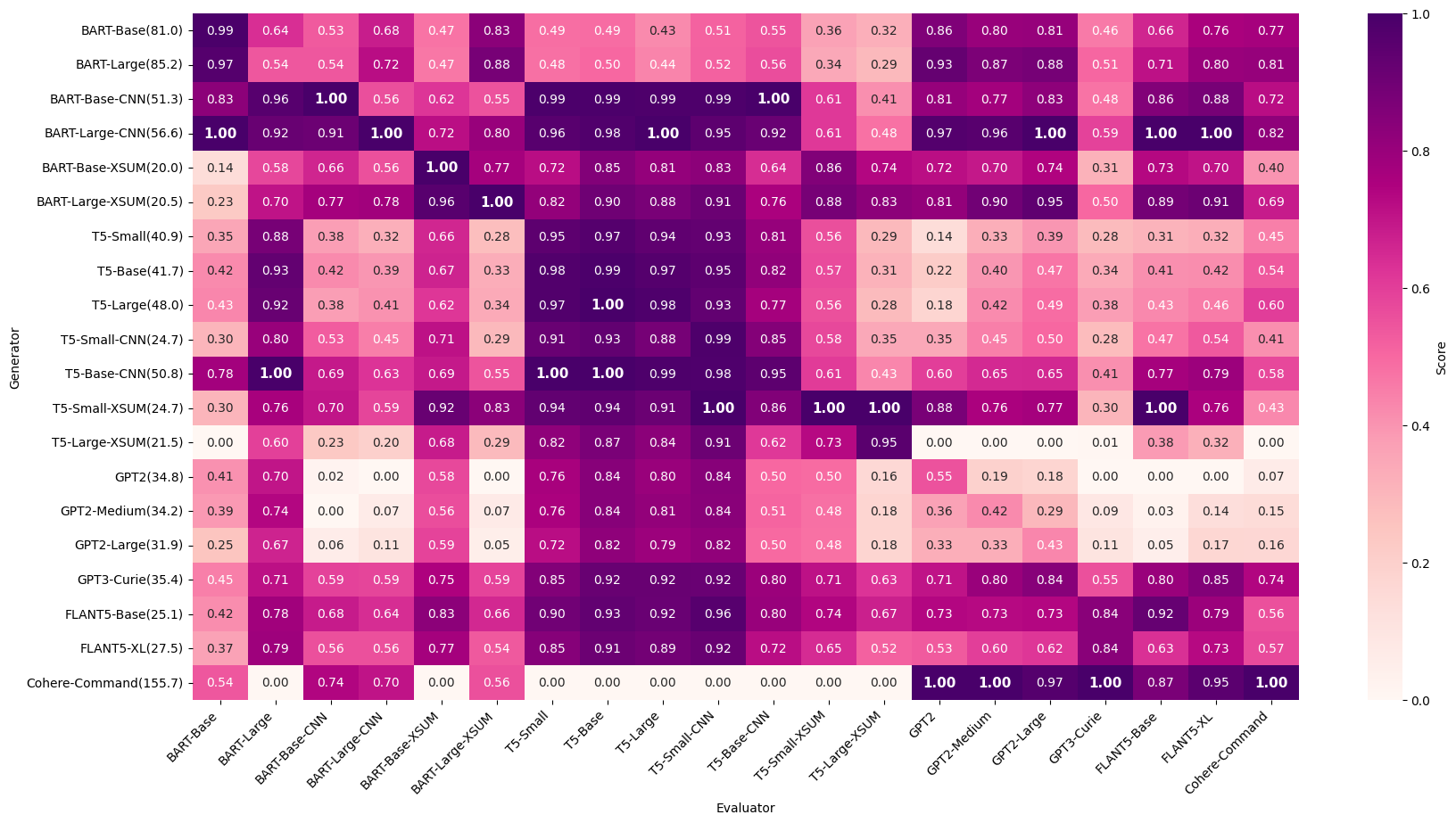}
\caption{Assessing Bias on CNN/DM Dataset using heatmaps in the \textit{reference-based} setting. Observing darker cells along the diagonal line indicates potential self-bias. All evaluator scores are normalized to a range between 0 and 1. Additionally, the number in the bracket represents the average length of summaries (measured in words) produced by the respective model.}
\label{Heat.cnn.f1}
\end{figure*}

% \begin{figure*}[!htbp]
% \centering  
% \subfigure[CNN/DM Dataset]{
% \label{Fig.sub.1}
% \includegraphics[width=.45\linewidth]{figures_AAAI/Length1.png}}\subfigure[XSUM Dataset]{
% \label{Fig.sub.2}
% \includegraphics[width=.45\linewidth]{figures_AAAI/Length2.png}}
% \caption{\textcolor{red}{Average Length of Summaries by Generators on Datasets CNN/DM and XSUM. -- duplicated info}}
% \textcolor{blue}{YL: think I can delete this Figure?}
% \label{LengthofSumm}
% \end{figure*}

Figures~\ref{Heat.cnn} and \ref{Heat.cnn.f1} display heatmaps presenting evaluator scores for various summaries generated by different generators from CNN/DM documents in reference-free and reference-base settings, respectively. These scores are computed by averaging the individual scores of the selected 500 documents. In both heatmaps, we observe darker cells along the diagonal line, running from the top left to the bottom right. This indicates the potential evaluator bias towards their corresponding generator models i.e., self-bias. However, this bias is notably more pronounced in the reference-free setting, commonly used for factuality evaluation \cite{koh-etal-2022-far}. 
%In this context, we highlight the presence of the darkest cells along the diagonal line.

Furthermore, as shown in Figure~\ref{Heat.cnn}, we note a distinct trend: T5-based generators, whether fine-tuned or not, tend to receive higher scores when assessed using different T5Score variations compared to evaluations using BARTScore,  GPTScore, or Cohere. This results in a concentrated dark rectangle at the heatmap's centre. 
%Notably, this pattern differs for T5 generators and evaluators that are fine-tuned on XSUM.
Similarly, we observe a parallel trend for BART-based generators, whether fine-tuned or not. 
%It is also quite evident that when using FLAN-T5 variants and Cohere as evaluators, they tend to favor the summaries generated by their respective models. 
%When evaluated \textcolor{blue}{by}
% with various 
%\textcolor{red}{
% \textcolor{blue}{identical} 
%variations \textcolor{blue}{with same evaluation frameworks}}, they consistently receive higher ranks compared to most of \textcolor{blue}{other} evaluations.
% while, evaluators are prone to assigning higher ranks to generators under the same dataset compared to those fine-tuned on others. For instance, when utilizing T5 fine-tuned on the XSUM dataset as evaluators, a preference is observed for BART-XSUM generators over T5-vanilla models.

\begin{figure*}[htbp]
\centering
\includegraphics[width=1\linewidth]{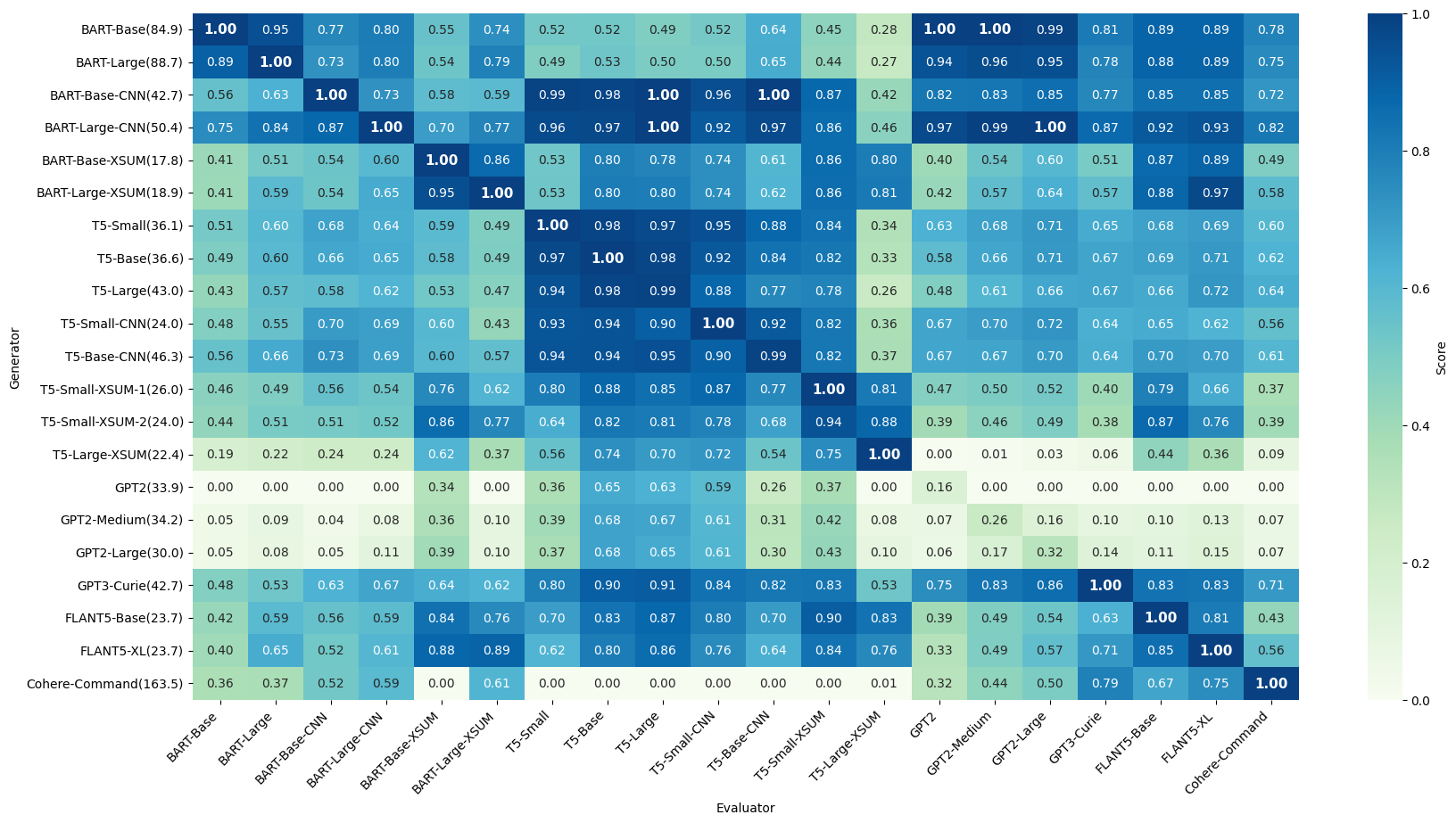}
\caption{Assessing Bias in the XSUM Dataset using heatmaps in the reference-free setting. Observing the darkest cells along the diagonal line, from the top left to the bottom right, indicates a distinct bias among evaluators towards their respective models. All evaluator scores are normalized to a range between 0 and 1. Additionally, the number in the bracket represents the average length of summaries (measured in words) produced by the respective model.}
\label{Heat.xsum}
\end{figure*}

Meanwhile, evaluators tend to assign higher ranks to generators trained on the same dataset as themselves, rather than to those fine-tuned on different datasets (see Figure~\ref{Heat.xsum}). For example, when using T5 models fine-tuned on the XSUM dataset as evaluators, there is a noticeable preference for BART-XSUM generators over T5-vanilla models, even though the evaluations are performed for the CNN Daily dataset. We observe the same pattern on summaries generated based XSUM documents. 

\begin{figure*}[!htb]
\centering
\includegraphics[width=.78\linewidth]{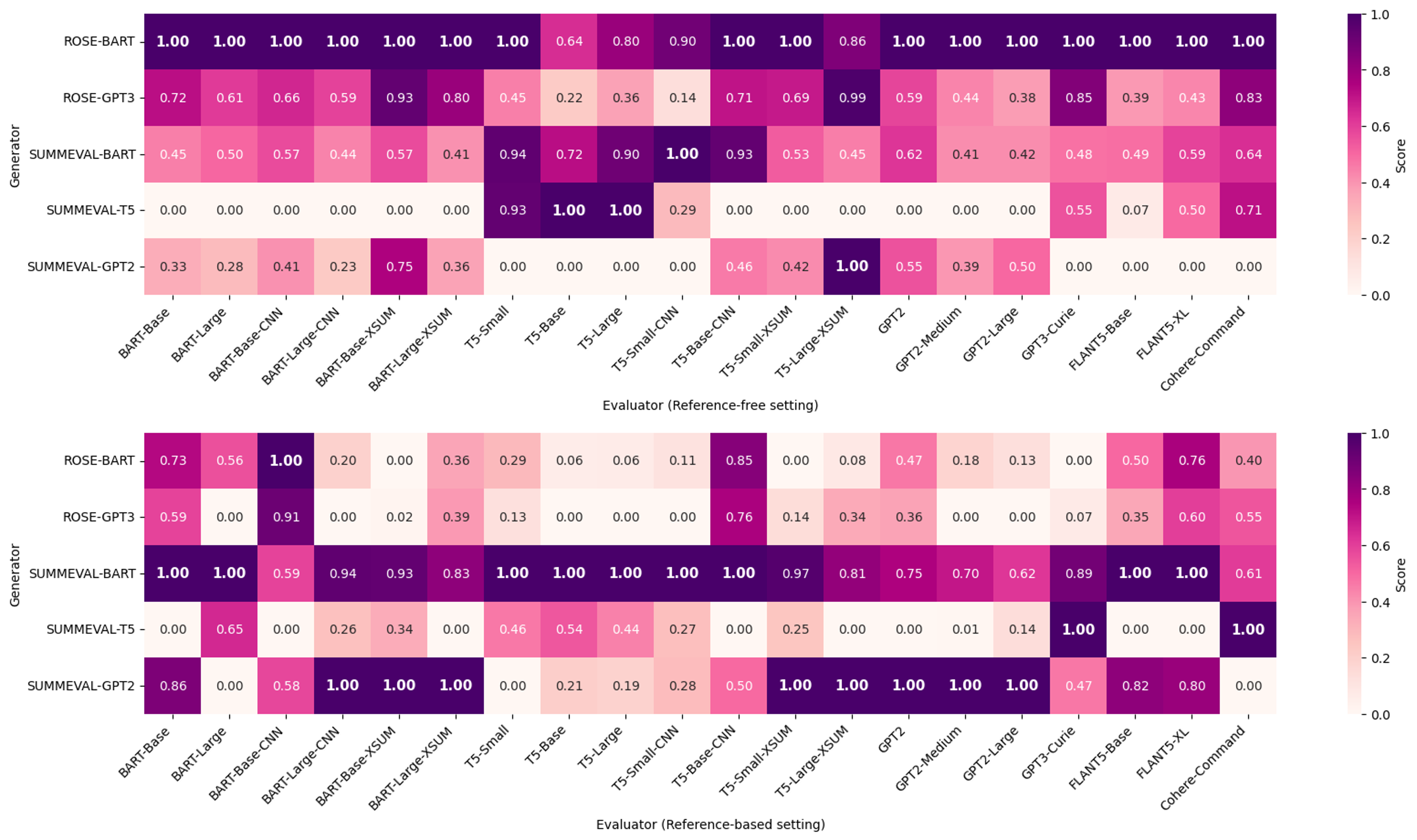}
\caption{Heatmaps of evaluation scores on the SummEval \& RoSE benchmarks for the reference-free and reference-based setting. We use the reference-free setting for SummEval and the reference-based setting for RoSE, aligning with the specific aspects each benchmark emphasizes.}
\label{RoseSummEval.Heatmap}
\end{figure*}

\begin{figure}[tb]
\centering
\includegraphics[width=.97\linewidth]{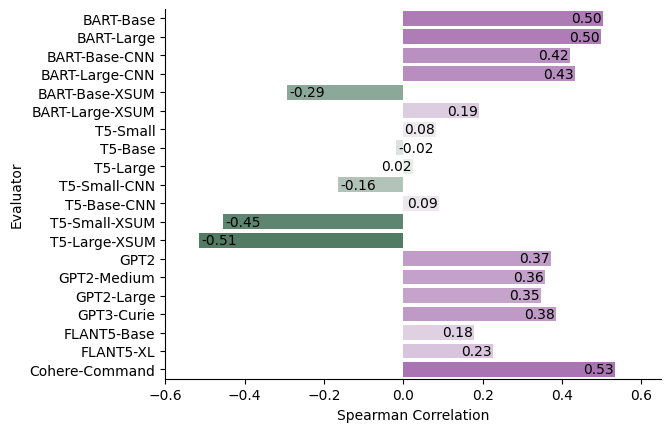}
\caption{Spearman Correlation between the length of generated summaries and the reference-free scores assigned by each evaluator. A higher positive score indicates that an evaluator prefers longer summaries, while a lower negative score indicates a preference for shorter summaries.}
\label{Leng.corr}
\end{figure}

\subsection{Bias towards Longer Summaries}
Another notable pattern in Figure~\ref{Heat.cnn} is the high scores for the BART-based generators, indicated by both BARTScore variants and different GPTScores. To further investigate this phenomenon, we calculate the average length of summaries generated by each generator for each of the datasets.  Notably, BART models and Cohere that have not been fine-tuned for summarization tend to produce the longest summaries on average. This is followed by the fine-tuned BART models on the CNN dataset. Conversely, T5-based models score the summaries generated by Cohere low, as they tend to favour shorter summaries. 
A similar preference for short summaries can also be observed for evaluators fine-tuned on XSUM, which one-sentence summaries.
%Similar observations can be observed that models fine-tuned on XSUM generate the shortest summaries.
%\footnote{Note that the XSUM dataset contains one-sentence summaries.}

Subsequently, we computed the Spearman correlation between the scores under the reference-free setting given by each of our examined evaluators and the length of the corresponding summary. The results are presented in Figure~\ref{Leng.corr}. Based on these results, with the exception of evaluators fine-tuned on XSUM, BARTScore and GPTScore variants tend to assign higher scores to longer summaries. This observation explains the darker squares positioned in the top-right corner of Figure~\ref{Heat.cnn} for high values of GPTScore variants, highlighting their inclination to assign higher scores to BART and BART-CNN generators that produce longer summaries. 
It is worth noting that this correlation with summary length is prominent within the reference-free setting. We observe a similar but less obvious pattern in the reference-based evaluations, as shown in Figure~\ref{Heat.cnn.f1}.
% \textcolor{blue}{but the right corner of Figure 3 is still a bit dark, isn't it? I think we should say this pattern is more prominent in Figure 2, not to say it is not true for Figure 3(\checkmark)}\textcolor{green}{Yes, I agree with this}

%\subsection{Length of Summary}

%Average length of summaries generated by each generator on dataset CNN/DM and XSUM is displayed respectively in Figure \ref{Length}. XSUM fine-tuned generators tend to produce concise summaries, while vanilla BART models on average generate longer content.

\subsection{Qualitative Analysis: Correlation of Self-Bias with Human Evaluation}
To further verify the evaluators’ self-bias, we repeat the experiments from \cref{unsupervised-experiments} on summarization benchmarks that are accompanied by human evaluations. While the number of summaries in these benchmarks is limited compared to those in \cref{unsupervised-experiments}, we can use the human annotations to verify that the inflated scores are not correlated with human evaluations.

%Both the SummEval and ROSE benchmarks include summaries that are scored very low by human evaluators. While automatic evaluators may agree on scoring low-quality summaries low, they may not necessarily be able to discriminate higher-quality summaries. As a result, we perform the experiments in this section under two different settings: (1) including all the summaries in the evaluation, and (2) including only the higher-quality summaries.

%We use the median score from human evaluators to select the higher-quality summaries. Specifically, we select summaries whose human scores are higher than the median value.\footnote{The distribution of human scores for each benchmark is shown in Table~\ref{tab:human_anno} in the Appendix.}

Figure \ref{RoseSummEval.Heatmap}
% \ref{Heat.Rosesummeval.ref-free} and Figure \ref{Heat.Rosesummeval.ref-based} 
shows the evaluation results for the SummEval and RoSE benchmarks for the reference-free and reference-based setting, respectively. As mentioned, we use  SummEval for the reference-free setting and  RoSE for the reference-based setting with regard to the specific aspects of each of these benchmarks \cite{Yuan2021-id}. 
% as aforementioned in section \cref{qualitative analysis}
% \footnote{The corresponding heatmap for the reference-based setting is included in the Appendix.}
% \textcolor{blue}{I think this is wrong, isn't it for both reference-free and reference-based. Similarly the caption of the figure is wrong. Shouldn't we then split this figure into two figures? Rose and summeval results wouldn't be comparable given they are based on different setting}\textcolor{green}{YL: Yes I agree. Since RoSE and SummEval uses different evaluation setting, they are not comparable. But if two graphs are splited and normalized seperately, there would only be 0 and 1 in RoSE part because there are only two rows, I was wondering if that would be alright.}
Overall, we observe a trend similar to that shown in Figure~\ref{Heat.cnn}. For instance, the T5-base generator receives higher scores from T5-based evaluators.\footnote{In SummEval, the T5 model is only ranked higher when evaluated with certain variants of the T5Score in the reference-based setting.} Meanwhile, BART-based models receive higher scores from both BARTScore and GPTScore evaluators, instead of T5 evaluator.
% \textcolor{blue}{the pattern doesn't seem to be that much compatible with Figure 2}

% For human correlations, we use the RoSE and SummEval benchmark for evaluating the correlation in the reference-based and reference-free setting respectively.
Table~\ref{tab:spear&kendal_reffree_summeval} presents the Spearman and Kendall correlation values of SummEval in the reference-free setting, whereas the Spearman and Kendall correlation values of RoSE in the reference-based setting are given in Table~\ref{tab:spear&kendal_refbase_rose}.

% \footnote{The reference-based values for RoSE are included in the Appendix.\ref{B}}
%The Pearson and Spearman correlations of each evaluator with the corresponding human evaluations in the ROSE and SummEval benchmarks are presented in Table~\ref{tab:pear_human} and Table~\ref{tab:spear_human}, respectively. 
Overall, we observe that none of the evaluators have a strong correlation with the human annotations on either of these benchmarks. Due to the limited size of the samples (i.e., 100 summaries from SummEval and 100 summaries from ROSE with human annotations, as described in \S\ref{dataset}) and the absence of many of our investigated generators in \cref{unsupervised-experiments}, we cannot draw a conclusive conclusion from the correlation values. Nevertheless, these results demonstrate that none of these evaluators highly correlate with human annotations, and as observed in \cref{unsupervised-experiments}, their inflated scores for their own underlying generator may contribute to this low correlation.

\input{tables/table_humancorr_refbase_rose}

\input{tables/table_humancorr_reffree_summeval}

% \input{tables/table_humancorr_refbase_rose}

% Furthermore, it is necessary to exclude the potential influences that the intrinsic quality of the generated content may exert on the score assignment. To achieve this, a subset of summaries that have been denoted as low quality by human annotators will be selected for evaluations. This will ascertain if the preference of evaluators still exist regardless of the intrinsic quality of generated content.

% To filter out lower-quality text, considering the median of each dataset's annotation, we remove summaries in RoSE with 'ACU' below 0.38, and SummEval entries with scores under 4.6. The result is shown in Figure \ref{Heat.ref-free.2}. Based on comparison of two figures: the bias of the evaluator becomes more pronounced when only high-quality summaries remain.

%Results from RoSE and SummEval are taken Meta-evaluation respectively due to different criteria are utilized by human annotations, assessed using Pearson and Spearman correlations. Table \ref{tab:pear_human} and Table \ref{tab:spear_human} illustrates Pearson and Spearman correlations on both dataset.

\section{Conclusions}
Based on experiments, we make the following conclusions:
\textbf{First}, the popularity of generative evaluation metrics, such as BARTScore, is on the rise for evaluating the factual accuracy of generated content—a critical concern in modern generator models. However, our results reveal that this evaluation approach is susceptible to the self-bias, highlighting the need for more robust metrics to assess factual correctness reliably.
\textbf{Second}, our analysis indicates that models fine-tuned on the XSUM dataset are not suitable for direct integration into evaluators due to their bias towards shorter summaries. The exception is their use for evaluating summaries aligned with XSUM-style content.
\textbf{Third}, notably, similar to traditional evaluation metrics \cite{sun-etal-2019-compare}, contemporary evaluation metrics might also lean towards favoring longer summaries. This bias should be considered when interpreting and applying these metrics.
\textbf{Finally}, our study uncovers the presence of the self-bias across all assessed evaluators. Consequently, we recommend avoiding the use of the same underlying model as the generator for assessment. Although the limited human evaluations for our examined models prevent definitive conclusions on selecting the best generative evaluator, our research charts a promising direction for designing more resilient and unbiased evaluation metrics.

In summary, our study identifies a new type of bias in generative evaluators encouraging future research in this direction for designing fairer evaluation metrics.

\section*{Limitations}
We note that our work has the following limitations. 
Firstly, our experiment has been focused on the summarization task. Expanding the evaluation to encompass a broader range of generation tasks would be highly beneficial. Secondly, conducting a larger-scale human evaluation would be advantageous, as our current experiments are constrained by the limited sample sizes from SummEval and RoSE. Finally, incorporating additional generation models and evaluators in future work would further enrich the experiment.

\section*{Ethics Statement}
This paper raises no ethical concerns. The data and supplementary materials  used in this study are open-sourced and widely employed in existing works.

\bibliography{custom}

\appendix

\section{Evaluation Setting}\label{A}

% \subsection{Human Annotated Benchmark}\label{A.1}

% \input{tables/table_humancorr_reffree_summeval}

\subsection{Generator}\label{A.1}
Full details of the models (e.g. checkpoint, prompt setting) that we employed as generators are given in Table \ref{tabs:name.gene}.
\input{tables/table_generators}

\subsection{Evaluator}\label{A.2}
Full details of the models that we employed as our evaluators are given in Table \ref{tabs:name.eval} (reference-free settings) and Table \ref{tabs:name.eval-refbased} (reference-based settings).
\input{tables/table_evaluators}
\input{tables/table_evaluators_refbased}

\section{Evaluation Results}\label{B}

\subsection{Reference-free Setting}\label{B.1}
% Heatmap of bias experiments on CNN/DM dataset in Reference-free setting is illustrated by Figure \ref{Heat.cnn}

Results of XSUM Dataset in Reference-free setting are presented in Figure \ref{Heat.xsum}. Evaluation scores for RoSE and SummEval benchmarks under the reference-free setting are shown in Figure \ref{RoseSummEval.Heatmap}.

For the meta evaluation, Spearman and Kendall correlation values in the reference-free setting for SummEval benchmark are shown in Table \ref{tab:spear&kendal_reffree_summeval}.

% \begin{figure*}[htbp]
% \centering
% \includegraphics[width=1\linewidth]{figures_AAAI/bias_heatmap_xsum_normalized.png}
% \caption{Assessing Bias in the XSUM Dataset using heatmaps in the reference-free setting. Observing the darkest cells along the diagonal line, from the top left to the bottom right, indicates a distinct bias among evaluators towards their respective models. All evaluator scores are normalized to a range between 0 and 1. Additionally, the number in the bracket represents the average length of summaries (measured in words) produced by the respective model.}
% \label{Heat.xsum}
% \end{figure*}

\subsection{Reference-based Setting}\label{B.2 refbased}
Heatmap of evaluation result on CNN/DM dataset under reference-based setting is given by Figure \ref{Heat.cnn.f1}

Evaluation scores for RoSE and SummEval benchmarks under the reference-based setting are illustrated by Figure \ref{RoseSummEval.Heatmap}.

% \begin{figure*}[htbp]
% \centering  
% \includegraphics[width=1\linewidth]{figures_AAAI/bias_heatmap_f1setting_rosesummeval.png}
% \caption{Heatmaps of evaluation scores on the ROSE \& SummEval benchmarks for the reference-based setting. It is recommended to utilize the reference-free setting for SummEval and the reference-based setting for ROSE, aligning with the specific aspects each benchmark emphasizes.}
% \label{Heat.Rosesummeval.ref-based}
% \end{figure*}

For the meta evaluation, Spearman and Kendall correlation values in the reference-based setting for RoSE benchmark are shown in Table \ref{tab:spear&kendal_refbase_rose}.

\end{document}

%% file: tables/table_annotation.tex
\begin{table}[tb] \small
\scalebox{1}{
\begin{tabular}{lllll}
\hline
              & Max  & Min  & Mean  & Median  \\ \hline
RoSE-BART     & 1.00      & 0.00      & 0.37       & 0.38         \\
RoSE-GPT3     & 0.90      & 0.00      & 0.27       & 0.25         \\ \hline
SummEval-BART & 5.00      & 2.67      & 4.57       & 4.67         \\
SummEval-T5   & 5.00      & 2.33      & 4.52       & 4.67         \\
SummEval-GPT2 & 5.00      & 1.33      & 3.57       & 3.58         \\ \hline
\end{tabular}}
\caption{
{Distribution of human annotation scores on the RoSE and SummEval datasets, where in RoSE we consider the ‘ACU’ score, and in SummEval we focus on four aspects—‘Coherence’, ‘Consistency’, ‘Fluency’, and ‘Relevance’—as evaluated by expert annotators. The scores for SummEval are obtained by averaging the scores across all aspects and evaluations from all annotators.}}
\label{tab:human_anno}
\end{table}

%% file: tables/table_humancorr_refbase_rose.tex
\begin{table}[!tb] \small
\begin{tabular}{lcc}
\hline
\multicolumn{3}{c}{\multirow{2}{*}{RoSE - Reference-based}} \\
\multicolumn{3}{l}{}                                                               \\ \hline
\multirow{3}{*}{Evaluator}         & \multicolumn{2}{c}{\multirow{2}{*}{ACU}}      \\
                                   & \multicolumn{2}{c}{}                          \\
                                   & Spearman              & Kendall               \\ \hline
BART-Base                          & 0.454                 & 0.310                 \\
BART-Large                         & 0.298                 & 0.218                 \\
BART-Base-CNN                      & \textbf{0.488}        & \textbf{0.345}        \\
BART-Large-CNN                     & 0.468                 & 0.329                 \\
BART-Base-XSUM                     & 0.150                 & 0.103                 \\
BART-Large-XSUM                    & 0.371                 & 0.253                 \\
T5-Small                           & 0.396                 & 0.284                 \\
T5-Base                            & 0.395                 & 0.285                 \\
T5-Large                           & 0.392                 & 0.282                 \\
T5-Small-CNN                       & 0.393                 & 0.281                 \\
T5-Base-CNN                        & 0.391                 & 0.276                 \\
T5-Small-XSUM                      & 0.379                 & 0.269                 \\
T5-Large-XSUM                      & 0.462                 & 0.324                 \\
GPT2                               & 0.375                 & 0.255                 \\
GPT2-Medium                        & 0.357                 & 0.244                 \\
GPT2-Large                         & 0.353                 & 0.242                 \\
GPT3-Curie                         & 0.310                 & 0.214                 \\
FLANT5-Base                        & 0.460                 & 0.325                 \\
FLANT5-XL                          & 0.433                 & 0.304                 \\
Cohere-Command                     & 0.384                 & 0.267                 \\ \hline
\end{tabular}

\caption{
{Spearman and Kendall correlations between reference-based evaluation scores and human annotations using annotations in RoSE. Results in bold indicate the strongest coefficient.
}}
\label{tab:spear&kendal_refbase_rose}

\end{table}

%% file: tables/table_humancorr_reffree_summeval.tex
\begin{table*}[!htb] \small
\begin{tabular}{lcccccccc}
\hline

\multicolumn{9}{c}{\multirow{2}{*}{SummEval - Reference-free}}                                                                                                                                                                 \\
\multicolumn{9}{c}{}                                                                                                                                                                                                           \\ \hline
\multirow{3}{*}{Evaluator} & \multicolumn{2}{c}{\multirow{2}{*}{Coherence}} & \multicolumn{2}{c}{\multirow{2}{*}{Consistency}} & \multicolumn{2}{c}{\multirow{2}{*}{Fluency}} & \multicolumn{2}{c}{\multirow{2}{*}{Relevance}} \\
                           & \multicolumn{2}{c}{}                           & \multicolumn{2}{c}{}                             & \multicolumn{2}{c}{}                         & \multicolumn{2}{c}{}                           \\
                           & Spearman               & Kendall               & Spearman                & Kendall                & Spearman              & Kendall              & Spearman               & Kendall               \\ \hline

BART-Base         & -0.028 & -0.021 & 0.107 & 0.078 & -0.043 & -0.037 & 0.105 & 0.074  \\
BART-Large        & 0.052  & 0.040  & 0.180 & 0.137 & 0.053  & 0.037  & 0.180 & 0.128  \\
BART-Base-CNN     & 0.193  & 0.138  & 0.228 & 0.171 & 0.190  & 0.145  & 0.069 & 0.050  \\
BART-Large-CNN    & 0.171          & 0.119          & 0.255          & 0.192      & 0.156         & 0.119        & 0.157   & 0.111  \\
BART-Base-XSUM    & 0.170          & 0.120          & -0.103         & -0.079     & 0.068         & 0.055        & -0.174  & -0.124 \\
BART-Large-XSUM   & 0.055          & 0.040          & 0.060                        & 0.046                       & -0.025                       & -0.022                      & 0.080                        & 0.056                       \\
T5-Small          & 0.208          & 0.146          & 0.547                        & 0.419                       & \textbf{0.501}               & \textbf{0.398}              & 0.415                        & 0.295                       \\
T5-Base           & 0.173          & 0.119          & 0.533                        & 0.409                       & 0.488                        & 0.381                       & 0.367                        & 0.260                       \\
T5-Large          & 0.185          & 0.132          & 0.477                        & 0.364                       & 0.445                        & 0.345                       & 0.387                        & 0.281                       \\
T5-Small-CNN      & \textbf{0.315} & \textbf{0.222} & 0.462                        & 0.356                       & 0.401                        & 0.314                       & 0.299                        & 0.214                       \\
T5-Base-CNN       & 0.192          & 0.135          & 0.253                        & 0.190                       & 0.189                        & 0.150                       & 0.148                        & 0.106                       \\
T5-Small-XSUM     & 0.245          & 0.178          & 0.142                        & 0.109                       & 0.209                        & 0.164                       & 0.113                        & 0.079                       \\
T5-Large-XSUM     & 0.213          & 0.152          & -0.111                       & -0.085                      & 0.018                        & 0.012                       & -0.041                       & -0.029                      \\
GPT2              & 0.103          & 0.077          & 0.154                        & 0.117                       & 0.037                        & 0.026                       & 0.032                        & 0.021                       \\
GPT2-Medium       & 0.123          & 0.091          & 0.234                        & 0.179                       & 0.117                        & 0.086                       & 0.066                        & 0.047                       \\
GPT2-Large        & 0.119          & 0.089          & 0.184                        & 0.140                       & 0.107                        & 0.080                       & 0.024                        & 0.017                       \\
GPT3-Curie        & 0.152          & 0.108          & 0.483                        & 0.371                       & 0.345                        & 0.264                       & 0.311                        & 0.223                       \\
FLANT5-Base       & 0.220          & 0.154          & 0.448                        & 0.345                       & 0.295                        & 0.228                       & 0.229                        & 0.159                       \\
FLANT5-XL         & 0.248          & 0.174          & \textbf{0.550}               & \textbf{0.424}              & 0.389                        & 0.301                       & 0.402                        & 0.289                       \\
Cohere-Command    & 0.136          & 0.097          & 0.520                        & 0.397                       & 0.351                        & 0.268                       & \textbf{0.427}               & \textbf{0.302}              \\ \hline
\end{tabular}

\caption{Spearman and Kendall correlations between the reference-free evaluation scores and expert annotations provided in SummEval on four different aspects. The strongest correlation for each aspect is bolded.}
\label{tab:spear&kendal_reffree_summeval}

\end{table*}

%% file: tables/table_generators.tex
\begin{table*}[!htb]\small
\begin{tabular}{llll}
\hline
Name of Generator & Name of Checkpoint or Model            & Suffix                      & Prefix                      \\ \hline
BART-Base         & facebook/bart-base                     & \XSolidBrush & Summarize:                  \\
BART-Large        & facebook/bart-large                    & \XSolidBrush & Summarize:                  \\
BART-Base-CNN     & ainize/bart-base-cnn                   & \XSolidBrush & \XSolidBrush \\
BART-Large-CNN    & facebook/bart-large-cnn                & \XSolidBrush & \XSolidBrush \\
BART-Base-XSUM    & morenolq/bart-base-xsum                & \XSolidBrush & \XSolidBrush \\
BART-Large-XSUM   & facebook/bart-large-xsum               & \XSolidBrush & \XSolidBrush \\
T5-Small          & t5-small                               & \XSolidBrush & Summarize:                  \\
T5-Base           & t5-base                                & \XSolidBrush & Summarize:                  \\
T5-Large          & t5-large                               & \XSolidBrush & Summarize:                  \\
T5-Small-CNN      & ubikpt/t5-small-finetuned-cnn          & \XSolidBrush & \XSolidBrush \\
T5-Base-CNN       & flax-community/t5-base-cnn-dm          & \XSolidBrush & \XSolidBrush \\
T5-Small-XSUM     & pki/t5-small-finetuned xsum            & \XSolidBrush & \XSolidBrush \\
T5-Large-XSUM     & sysresearch101/t5-large-finetuned-xsum & \XSolidBrush & \XSolidBrush \\
GPT2              & openai-community/gpt2                  & TL;DR:                      & \XSolidBrush \\
GPT2-Medium       & openai-community/gpt2-medium           & TL;DR:                      & \XSolidBrush \\
GPT2-Large        & openai-community/gpt2-large            & TL;DR:                      & \XSolidBrush \\
GPT3-Curie        & text-curie-001                         & TL;DR:                      & \XSolidBrush \\
FLANT5-Base       & google/flan-t5-base                    & TL;DR:                      & \XSolidBrush \\
FLANT5-XL         & google/flan-t5-xl                      & TL;DR:                      & \XSolidBrush \\
Cohere-Command    & api.cohere.ai/v1/generate              & \XSolidBrush                & Write a concise summarization: \\  \hline
\end{tabular}
\caption{Checkpoints or model utilized in our generation setting with corresponding prompt configurations, ‘text-curie-001’ is the model name provided by OpenAI API, and ‘api.cohere.ai/v1/generate’ denotes model names provided by Cohere API, alongside other checkpoints available through Hugging Face.}
\label{tabs:name.gene}
\end{table*}

%% file: tables/table_evaluators.tex
\begin{table*}[!htb] \small
\begin{tabular}{llll}
\hline
Name of Evaluator & Name of Checkpoint or Model            & Suffix                      & Prefix                      \\ \hline
BART-Base         & facebook/bart-base                     & \XSolidBrush & Summarize:                  \\
BART-Large        & facebook/bart-large                    & \XSolidBrush & Summarize:                  \\
BART-Base-CNN     & ainize/bart-base-cnn                   & \XSolidBrush & \XSolidBrush \\
BART-Large-CNN    & facebook/bart-large-cnn                & \XSolidBrush & \XSolidBrush \\
BART-Base-XSUM    & morenolq/bart-base-xsum                & \XSolidBrush & \XSolidBrush \\
BART-Large-XSUM   & facebook/bart-large-xsum               & \XSolidBrush & \XSolidBrush \\
T5-Small          & t5-small                               & \XSolidBrush & Summarize:                  \\
T5-Base           & t5-base                                & \XSolidBrush & Summarize:                  \\
T5-Large          & t5-large                               & \XSolidBrush & Summarize:                  \\
T5-Small-CNN      & ubikpt/t5-small-finetuned-cnn          & \XSolidBrush & \XSolidBrush \\
T5-Base-CNN       & flax-community/t5-base-cnn-dm          & \XSolidBrush & \XSolidBrush \\
T5-Small-XSUM     & pki/t5-small-finetuned xsum            & \XSolidBrush & \XSolidBrush \\
T5-Large-XSUM     & sysresearch101/t5-large-finetuned-xsum & \XSolidBrush & \XSolidBrush \\
GPT2              & openai-community/gpt2                  & TL;DR:                      & \XSolidBrush \\
GPT2-Medium       & openai-community/gpt2-medium           & TL;DR:                      & \XSolidBrush \\
GPT2-Large        & openai-community/gpt2-large            & TL;DR:                      & \XSolidBrush \\
GPT3-Curie        & text-curie-001                         & TL;DR:                      & \XSolidBrush \\
FLANT5-Base       & google/flan-t5-base                    & TL;DR:                      & \XSolidBrush \\
FLANT5-XL         & google/flan-t5-xl                      & TL;DR:                      & \XSolidBrush \\
Cohere-Command    & api.cohere.ai/v1/generate              & \XSolidBrush                & Write a concise summarization: \\  \hline
\end{tabular}
\caption{Checkpoints or model utilized in our evaluation study for the reference-free setting with corresponding prompt configurations, ‘text-curie-001’ is the model name provided by OpenAI API, and ‘api.cohere.ai/v1/generate’ denotes model names provided by Cohere API, alongside other checkpoints available through Hugging Face.}
\label{tabs:name.eval}
\end{table*}

%% file: tables/table_evaluators_refbased.tex
\begin{table*}[!htb] \small
\begin{tabular}{llll}
\hline
Name of Evaluator & Name of Checkpoint or Model            & Suffix          & Prefix                       \\ \hline
BART-Base         & facebook/bart-base                     & in other words: & \XSolidBrush                 \\
BART-Large        & facebook/bart-large                    & in other words: & \XSolidBrush                 \\
BART-Base-CNN     & ainize/bart-base-cnn                   & \XSolidBrush & \XSolidBrush \\
BART-Large-CNN    & facebook/bart-large-cnn                & \XSolidBrush & \XSolidBrush \\
BART-Base-XSUM    & morenolq/bart-base-xsum                & \XSolidBrush & \XSolidBrush \\
BART-Large-XSUM   & facebook/bart-large-xsum               & \XSolidBrush & \XSolidBrush \\
T5-Small          & t5-small                               & \XSolidBrush & Paraphrase:  \\
T5-Base           & t5-base                                & \XSolidBrush & Paraphrase:  \\
T5-Large          & t5-large                               & \XSolidBrush & Paraphrase:  \\
T5-Small-CNN      & ubikpt/t5-small-finetuned-cnn          & \XSolidBrush & \XSolidBrush \\
T5-Base-CNN       & flax-community/t5-base-cnn-dm          & \XSolidBrush & \XSolidBrush \\
T5-Small-XSUM     & pki/t5-small-finetuned xsum            & \XSolidBrush & \XSolidBrush \\
T5-Large-XSUM     & sysresearch101/t5-large-finetuned-xsum & \XSolidBrush & \XSolidBrush \\
GPT2              & openai-community/gpt2                  & Paraphrase the sentence:    & \XSolidBrush \\
GPT2-Medium       & openai-community/gpt2-medium           & Paraphrase the sentence:    & \XSolidBrush \\
GPT2-Large        & openai-community/gpt2-large            & Paraphrase the sentence:    & \XSolidBrush \\
GPT3-Curie        & text-curie-001                         & Paraphrase the sentence:    & \XSolidBrush \\
FLANT5-Base       & google/flan-t5-base                    & Paraphrase the sentence:    & \XSolidBrush \\
FLANT5-XL         & google/flan-t5-xl                      & Paraphrase the sentence:    & \XSolidBrush \\
Cohere-Command    & api.cohere.ai/v1/generate              & Paraphrase the sentence:    & \XSolidBrush \\  \hline
\end{tabular}
\caption{Checkpoints and models utilised in our evaluation study for the reference-based setting with corresponding prompt configurations, ‘text-curie-001’ is the model name provided by OpenAI API, and ‘api.cohere.ai/v1/generate’ denotes model names provided by Cohere API, alongside other checkpoints available through Hugging Face.}
\label{tabs:name.eval-refbased}
\end{table*}